\documentclass[10pt,twocolumn,letterpaper]{article}

\usepackage[pagenumbers]{cvpr}
\usepackage{graphicx}
\usepackage{amsmath}
\usepackage{amssymb}
\usepackage{mathtools}
\usepackage{booktabs}
\DeclareMathOperator*{\argmax}{arg\,max}
\usepackage[belowskip=-1pt,aboveskip=0pt]{caption}
\usepackage{enumitem}
\usepackage{multirow}
\usepackage{pifont}
\usepackage{dsfont}
\usepackage[utf8]{inputenc} %
\usepackage[T1]{fontenc}    %
\usepackage{url}            %
\usepackage{booktabs}       %
\usepackage{amsfonts}       %
\usepackage{nicefrac}       %
\usepackage{microtype}      %
\usepackage{xcolor}         %
\usepackage{bbm}
\usepackage[pagebackref,breaklinks,colorlinks]{hyperref}
\usepackage[capitalize]{cleveref}
\crefname{section}{Sec.}{Secs.}
\Crefname{section}{Section}{Sections}
\Crefname{table}{Table}{Tables}
\crefname{table}{Tab.}{Tabs.}

\newcommand{\cmark}{\ding{51}}%

\newcommand{\fullname}{Knowledge-Guided Simple Primitives}
\newcommand{\fullnameBold}{\textbf{K}nowledge-guided \textbf{S}imple \textbf{P}rimitives}
\newcommand{\nickMethod}{\nickMethodNS\ }
\newcommand{\nickMethodNS}{KG-SP}
\newcommand{\settingNick}{\settingNickNS\ }
\newcommand{\settingNickNS}{pCZSL}

\newcommand{\myparagraph}[1]{\vspace{1pt}\noindent{\bf #1}}

\def\cvprPaperID{8438} %
\def\confName{CVPR}
\def\confYear{2022}

\begin{document}

\title{KG-SP: Knowledge Guided Simple Primitives \\ for Open World Compositional Zero-Shot Learning}

\author{Shyamgopal Karthik$^{1}$
~~~~~~~~~
Massimiliano Mancini$^{1}$
~~~~~~~~~
Zeynep Akata$^{1,2}$
~~~~~~~~~
~\\
\small{
$^1$University of T\"{u}bingen\quad\quad} 
\small{
$^2$Max Planck Institute for Intelligent Systems\quad\quad
}
}
\maketitle

\begin{abstract}
   The goal of open-world compositional zero-shot learning (OW-CZSL) is to recognize compositions of state and objects in images, given only a subset of them during training and no prior on the unseen compositions. %
   In this setting, models operate on a huge output space, containing all possible state-object compositions. While previous works tackle the problem by learning embeddings for the compositions jointly, %
    here we revisit a simple CZSL baseline and predict the primitives, i.e. states and objects, independently. %
   To ensure that the model develops primitive-specific features, we equip the state and object classifiers with separate, 
   non-linear feature extractors. Moreover, %
   we estimate the feasibility of each composition %
   through external knowledge, using this prior to remove unfeasible compositions from the output space.  
   Finally, we propose a new setting, i.e. {CZSL} under partial supervision (\settingNickNS), where either only objects or state labels are available during training, and we can use our prior to estimate the missing labels. Our model, \fullname\ (\nickMethodNS), %
     achieves state of the art in both OW-CZSL and \settingNickNS, 
     surpassing most recent competitors even when coupled with semi-supervised learning techniques. Code available at: \url{https://github.com/ExplainableML/KG-SP}.%

\end{abstract}

\section{Introduction}
\label{sec:intro}
As humans, we interact with objects depending on their state. For instance, we use ripe lemons rather than moldy ones to prepare a lemonade, and we clean dirty dishes after using them. Algorithms that can recognize objects together with their state are crucial for autonomous agents to show the same high-level interactions capabilities we have.
In the literature, this problem is studied under the name of \textit{Compositional Zero-shot Learning} (CZSL). In CZSL, we are given a training set with images of objects in a subset of their possible states and, at test time, the goal is to recognize compositions of the same set of objects and states, even unseen during training.  Since an object has a different appearance depending on its state (\eg \textit{dry dog} vs \textit{wet dog}) and a state modifies objects in different ways (\eg \textit{wet dog} vs \textit{wet car}), the challenge of CZSL is modeling how states and objects interact with each other, extrapolating this knowledge from seen to unseen compositions. Under this perspective, multiple works modeled the interactions of objects and states, either through compositional classifiers \cite{redwine,tmn}, or a shared embedding space \cite{aopp,symnet,cge}.

 \begin{figure}[t]
     \centering
     \includegraphics[width=\linewidth]{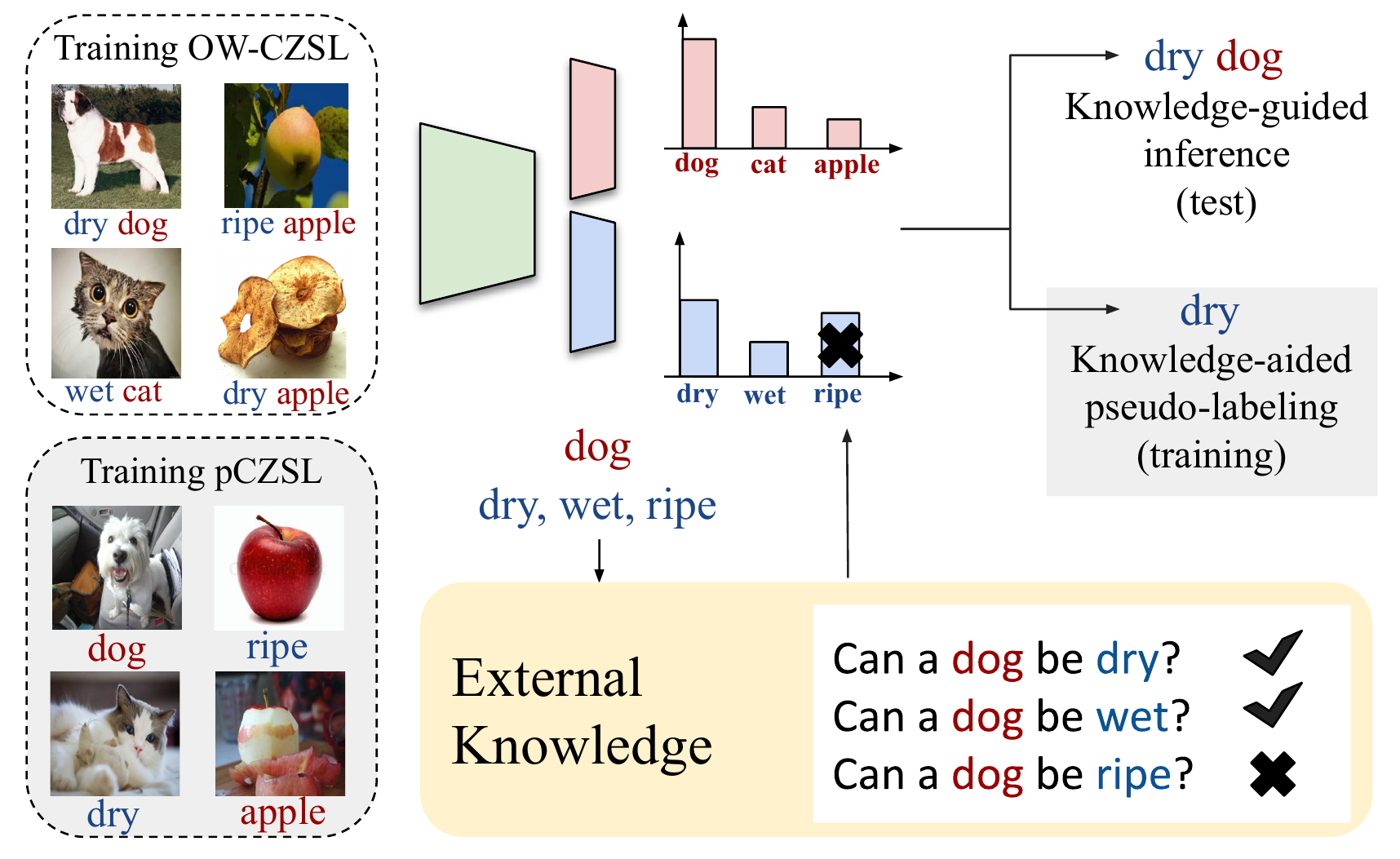} %
     \caption{We consider the problems of open-world CZSL (OW-CZSL), where we lack priors on unseen compositions at test time, and CZSL under partial supervision (pCZSL) where we also lack compositional labels during training (left). We tackle them by independently predicting object (red) and state (blue) labels %
     and by using external knowledge (bottom) %
     to estimate the feasibility of compositions, reducing the search space during inference and improving pseudo-labeling during training in pCZSL. %
     }
     \vspace{-10pt}
     \label{fig:teaser}
 \end{figure}

Despite their effectiveness, \cite{compcos} showed how the performance of CZSL methods degrade in the open-world setting (OW-CZSL). In OW-CZSL, there are no priors on the unseen compositions, and models must consider all possible compositions at test time. Due to the large cardinality of the output space, it is difficult to produce discriminative embeddings for the unseen compositions \cite{compcos}. Inspired by the findings of \cite{compcos}, in this work we explore a completely different direction. Specifically, we design an architecture that disregards the compositional nature of the problem and produces the initial predictions independently for objects and states. The idea is that while discriminating between compositions is hard in OW-CZSL due to the large search space, recognizing primitives (\ie objects and states) in isolation is easier since 1) the cardinality of the two sets is much lower and 2) the sets are fixed at both training and test time. %

Inspired from %
\cite{redwine} and \cite{DAP}, we design a simple method that 
predicts objects and states with two independent classifiers. %
Since recognizing states requires different features \wrt recognizing objects, instead of having a %
shared feature representation, we train our model with two different non-linear feature extractors. %
Furthermore, since not all compositions are equally feasible in reality (\eg \textit{ripe dog}) %
we can refine the predictions of our model by eliminating less feasible compositions from the output space. With this goal, we use external knowledge (\ie ConceptNet \cite{conceptnet}) to estimate the compatibility between a state and an object, using these estimates to remove less feasible compositions at test time. We name our model \fullname\ (\nickMethodNS). %
As our \nickMethod method does not require compositional labels during training,  we explore %
a new %
challenging setting, \ie CZSL under partial supervision (\settingNickNS). In \settingNickNS, training samples have either only object or state annotation, but not both. Here we use our prior on feasible compositions to aid pseudo-labeling during training. 
Experiments show that \nickMethod is either competitive or surpasses the current state of the art in OW-CZSL and outperforms recent CZSL approaches on \settingNick setting. Figure~\ref{fig:teaser} provides an overview of \nickMethod and the two tasks. %

\myparagraph{Contributions.} \noindent To summarize, %
    1) inspired by \cite{redwine,DAP}, our model predicts state and objects independently %
    while at the same time removing less feasible compositions from the output space based on external contextual information about the feasibility of certain compositions; 
    2) we explore the problem of CZSL under partial supervision, where either object or state information is missing in the ground-truth;
    3) we adapt recent baselines for \settingNick %
    showing that \nickMethod outperforms them even when coupled with semi-supervised learning techniques in both OW-CZSL and \settingNick settings. %

\section{Related works}
\label{sec:related}

\myparagraph{Compositional zero-shot learning} aims to recognize compositions of states and objects in images, even unseen during training. The main challenge of this setting is modeling how states modify objects, generalizing this capability to unseen compositions. Most of the previous works focused on how to model the interactions between states and objects either at the parameter level or in a given representation space. For instance, \cite{redwine,tmn} proposed to generate a classifier for a given state-object composition given two classifiers (or embeddings) for specific state and object primitives, using either a compositional module \cite{redwine} or a gating network \cite{tmn}. %
Differently, \cite{aopp,symnet} model each state as an operator transforming object embeddings, imposing properties on the state operators (\eg commutativity, symmetry). In \cite{aopp} the state operators are linear, while in \cite{symnet} they are coupling and decoupling networks. %
Recently, \cite{cge,ruis2021protoprop} used  graph convolutional networks~\cite{gcn} to model the interactions between state, objects and their compositions. %
Differently, \cite{causal} tackles CZSL from a causality perspective, learning disentangled objects and states representations. 
In this work, we {revisit VisProd  \cite{redwine}, predicting objects and states in isolation, }%
showing that {this strategy is effective in }%
OW-CZSL. {As in \cite{compcos}, we estimate the
feasibility of each composition to improve the model’s performance. However, we use ConceptNet to this aim, rather than compositional annotations, being the first to tackle %
 CZSL %
 without compositional labels during training.} %

\myparagraph{Multi-task learning.} %
Since we predict state and object independently, our work is related to Multi-Task \cite{caruana1997multitask,kendall2018multi,misra2016cross,rusu2016progressive,ruder2019latent} and Multi-Domain learning \cite{rebuffi2017learning,rebuffi2018efficient,li2019efficient,berriel2019budget}, where the goal is to learn a unique model able to address different visual tasks. Most of the approaches in this domain either learn task-specific parameters \cite{rebuffi2017learning,rebuffi2018efficient,li2019efficient,berriel2019budget, rusu2016progressive} and how to combine them \cite{misra2016cross,ruder2019latent}, or focus on re-weighting different loss functions \cite{kendall2018multi}. %
While we use multi-task learning to design primitive classifiers for CZSL, %
our final goal is different since we compose predictions from separate output spaces. %

\myparagraph{Learning from Partial Supervision.} Our CZSL setting without compositional labels %
is related to semi-supervised learning and learning with missing labels.  In semi-supervised learning, both labeled and unlabeled samples are available, and the goal %
is to effectively use the unlabeled samples. Popular ideas revolve around consistency regularization~\cite{berthelot2019mixmatch,berthelot2019remixmatch,sohn2020fixmatch}, %
and self-training~\cite{grandvalet2005semi,lee2013pseudo,rizve2021defense}. %
Unlike semi-supervised learning, in \settingNickNS, all samples are labeled, but partially. %
Thus, we can also exploit the prior on how objects interact with states to  estimate the missing labels. %

For what concerns learning with missing labels, 
this is most prevalent in multi-label scenarios where it is unfeasible to annotate all labels that are present in a single image. Approaches in this field usually model the correlation among labels \cite{durand2019learning,huynh2020interactive,kundu2020exploiting,cole2021multi} %
to impose semantic objectives on missing ones. %
While we are also interested in learning from partial supervision, %
our labels lie on two separate spaces (\ie objects and states), and the missing labels (\eg state) influences the appearance of the positive one (\eg object). In this setting, the main challenge is to model how the two spaces influence each other without any compositional supervision.

\section{Knowledge Guided Simple Primitives}
\label{sec:method}
\vspace{-5pt}

\myparagraph{Problem formulation.} CZSL \cite{compcos} aims to recognize compositions of a set of objects $O$ and a set of states $S$. %
Formally, we are given a training set $\mathcal{T} = \{(x,y)\}_{i=1}^N$, where $N$ is the size of the training set, %
$x\in X$ denotes an image in the input 
space $X$ and $y\in Y_s$ is its label in the set of seen compositions $Y_s$. The goal is to learn a model that can recognize a set of compositions $Y_t=Y_s\cup Y_u$, where $Y_u$ is a set of unseen compositions (\ie $Y_u\cap Y_s = \emptyset$) and $Y_t\subseteq Y$, with $Y$ being the set of all possible compositions, \ie $Y=S\times O$. %

\myparagraph{OW-CZSL and \settingNick settings.}
In this work, we consider two different CZSL settings. %
Open-World CZSL (OW-CZSL) \cite{compcos} %
assumes no prior on the set of unseen compositions %
at test time. This means that the model needs to operate on the full compositional space, \ie $Y_t=Y$. Consequently, the number of unseen compositions is much larger than the number of seen ones \ie $Y_u=Y\setminus Y_s$, thus the main challenge is operating in a very large output space where most of the compositions are unseen and thus hard to discriminate. %

In this work, %
we also consider a new challenging task, namely CZSL under %
partial supervision (\settingNickNS), where the training set does not contain any compositional label and all training images have either object or state label, but not both. This setting is more realistic than standard CZSL since most datasets are collected with single labels (\eg only object-level information) and collecting multiple labels is expensive and time-consuming.
Formally, %
we consider the labels of our training set $\mathcal{T}$ to be of the form $y=(s,\mathtt{u}) \vee  y=(\mathtt{u},o), \;\forall (x,y)\in\mathcal{T}$, with $s\in S$, $o\in O$ and $\mathtt{u}$ denoting an unknown label.  %
Note that, as a consequence of this formulation, the set of training compositions $Y_s$ is not known a priori anymore. This implies that, as in OW-CZSL, we need to consider the full compositional space at test time, \ie $Y_t=Y$. %
Moreover, since no training image contains both object and state labels, we do not have explicit supervision on how states modify objects and vice-versa. 

In the following, we describe the two components of our framework, Simple Primitives (SP) where we predict the primitives, e.g. object and states, independently and Knowledge Guidance (KG) where we use external resources that guide our model on the feasibility of certain compositions.

\subsection{Simple Primitives (SP) in KG-SP}

Inspired by the early Visual Product (VisProd) baseline~\cite{redwine}, our model completely disregards the compositional nature of the problem and predicts states and objects independently. %
This idea contrasts with %
 recent approaches (\ie \cite{aopp,tmn,symnet,cge,causal,compcos}), explicitly modeling the interactions between objects and states within the model. %

Formally, given an image $x$, we extract its feature representation %
$z = \omega(x)$ through a function $\omega$, mapping images into a feature space $Z$, \ie $\omega: X \rightarrow Z$. We then have an object classifier $\phi_o: Z \rightarrow \Delta_O$ that maps $z$ to a vector in the probability simplex $\Delta_O$, spanning all object categories. %
Similarly, we have another classifier that maps $z$ to a probability over the states, \ie %
$\phi_s: Z\rightarrow \Delta_S$. %
During training, we minimize the %
cross-entropy loss %
for both the object and state predictions. Specifically, we minimize:
\begin{align}
\label{eq:loss-visprod}
    \mathcal{L}_{\text{visprod}} &= \sum_{i=1}^N \mathbb{I}_{s_i\neq \mathtt{u}}\mathcal{L}_\text{state}(x_i,s_i) + \mathbb{I}_{o_i\neq \mathtt{u}} \mathcal{L}_\text{obj}(x_i,o_i)  \\
    &=  -\sum_{i=1}^N  \mathbb{I}_{s_i\neq \mathtt{u}}\log\phi_s(z_i,s_i) +\mathbb{I}_{o_i\neq \mathtt{u}}\log \phi_o(z_i,o_i) \nonumber
\end{align}
where $z_i=\omega(x_i)$, $\phi_o(z,o)$ is the probability of the object $o$ assigned by $\phi_o$ to the input $z$, and $\phi_s(z,s)$ is the probability of the state $s$ assigned by $\phi_s$ to the input $z$. In Eq.~\eqref{eq:loss-visprod}, $\mathbb{I}$ is an indicator function used to not compute the loss in \settingNickNS, in the absence of primitive labels. Our prediction function is:
\begin{equation} \label{eq:inf}
    f = \argmax_{(s,o)\in Y}\,\phi_o(w(x),o)\cdot \phi_s(w(x),s).
\end{equation} 
Although learning simple primitives independently like this may not be effective in standard CZSL, %
we argue that the capability to separate state and objects predictions is crucial in OW-CZSL, where the search space is too large if predictions are made over the full compositional space. %

\begin{figure*}[t]
    \centering
    \includegraphics[width=1.0\linewidth]{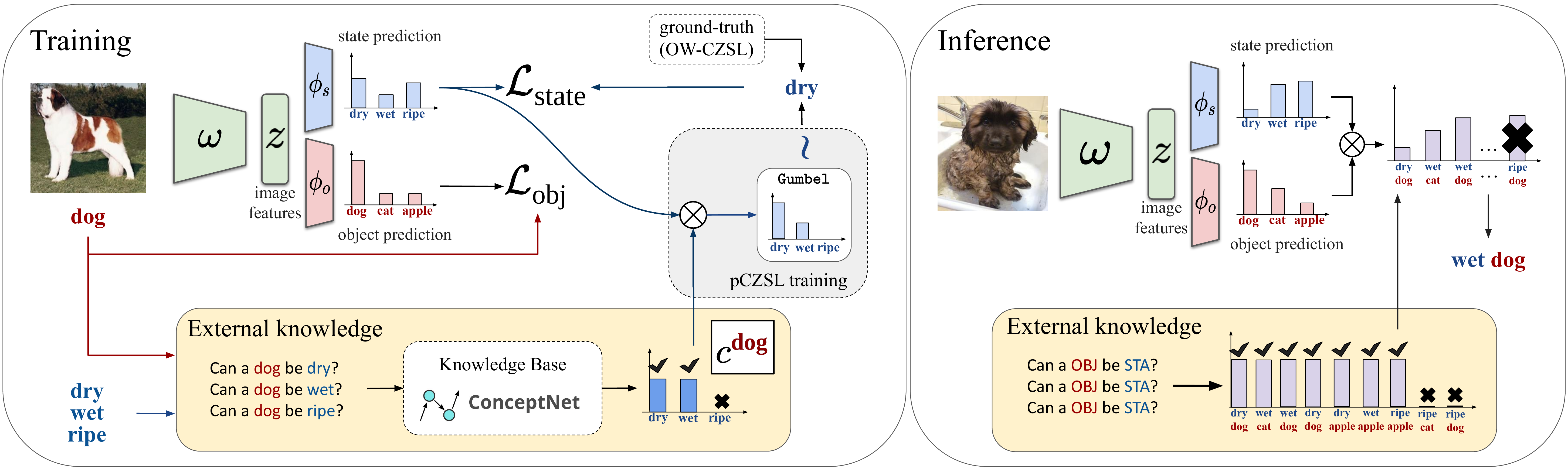}\vspace{5pt}
    \caption{\textbf{\fullname\ (\nickMethodNS)}. We train a separate object (red) and state (blue) predictor on top of a shared feature extractor (green) using the available state and object labels. We use external knowledge to estimate feasibility scores of compositions (yellow, bottom), using this prior during inference to directly remove unfeasible compositions from the output space. In \settingNick, we %
    use this knowledge to re-weight the class scores and perform pseudo-labeling (grey) of missing labels, sampling them through the Gumbel-softmax.} %
    \label{fig:model-fig}
    \vspace{-8pt}
\end{figure*}

In the original VisProd formulation, objects and state predictors are simple linear layers operating over %
the same feature vector. %
However, this choice is suboptimal and leads to low results in practice. %
In fact, by using separate linear layers, VisProd addresses CZSL as a multi-task learning problem \cite{caruana1997multitask,kendall2018multi,rebuffi2017learning,rebuffi2018efficient,misra2016cross,rusu2016progressive,ruder2019latent}, where there are two different tasks (\ie states and objects prediction) that share the same feature extractor while differing only for the classification head. However, multiple works in multi-task (MTL) and multi-domain learning (MDL) discussed how fully sharing the parameters to extract the feature representation for different tasks (\ie \textit{hard-sharing} \cite{caruana1997multitask}) %
is suboptimal when the tasks are not strictly related %
\cite{rebuffi2017learning,rebuffi2018efficient,misra2016cross} and may even lead to negative transfer \cite{lee2018deep}. %

In CZSL, recognizing objects is different than recognizing their states. Specifically, the former requires focusing on global features: for instance, distinguishing an animal from another requires focusing on their shapes and skins while distinguishing fruits requires detecting texture-based cues. On the contrary, recognizing states requires focusing on local patterns: for instance, the difference between \textit{dry} and \textit{wet} can be detected by the presence of drops in \textit{cars} and \textit{apples} while in animals it requires looking at the shape of the fur. With this premise, we need to overcome the limits imposed by hard-parameter sharing to ensure the objects and state classifiers have enough flexibility to learn primitive-specific features. While advanced MTL and MDL techniques can be used for this purpose, in this work we found that it is sufficient to implement the two classifiers as multi-layer perceptrons (MLP) with non-linear activations. %

\subsection{Knowledge Guidance (KG) in KG-SP}

In the large output space of OW-CZSL and \settingNickNS, not all compositions are equally feasible (\eg \textit{ripe dog}, \textit{hairy apple}) and taking this prior into account can help in correcting incompatible state-object predictions of our model. %
In the following we describe how we estimate the feasibility scores and how we use them in our model. 

\myparagraph{Estimating feasibility scores.}
Formally, let us associate to %
each composition $(s,o)$ a compatibility score $c_{s}^o\in[0,1]$. %
Since there exist no database contain such information, %
previous works exploited the set of seen compositions $Y_s$ to estimate $c_s^o$\cite{compcos}.
Here we explore an alternative direction by using external knowledge. In this way, our estimation is independent of the actual availability of the set $Y_s$ and can also be applied in \settingNickNS, where $Y_s$ is unknown. While we explored different strategies (see supplementary), we found ConceptNet \cite{conceptnet} to give reliable feasibility estimates. %

ConceptNet is a {knowledge graph connecting words and phrases with labeled edges, extracted from various sources \cite{conceptnet}.} We can use ConceptNet in two ways. The first is querying for the existence of a composition and the second is querying for the relatedness between two entries (\ie object and state). Since direct queries are very sparse, we follow the second approach, defining the scores as: %
\begin{equation}
 \label{eq:kb-relation}
     c_{s}^o = \rho_\text{KB}(s,o)
\end{equation}
where $\rho_\text{KB}(s,o)$ returns the relational score between $s$ and $o$. In ConceptNet, these scores are computed from the cosine-similarity of \textit{ConceptNet Numberbatch} embeddings \cite{speer2017conceptnetemb}. {The latter are built from ConceptNet adjacency matrix and existing word embeddings (e.g. word2vec \cite{word2vec}, GloVe~\cite{pennington2014glove}).}

\myparagraph{Using the feasibility scores during inference.} Similarly to \cite{compcos}, the most straightforward-way to use the feasibility sores is to 
remove from the output space less feasible compositions during inference. Thus, our prediction function becomes:
\begin{equation} 
\label{eq:inf-hm}
    f_{\text{KG}} = \argmax_{(s,o)\in Y, c_s^o>0}\,\phi_o(z,o)\cdot \phi_s(z,s)\end{equation} 
where we consider feasible all compositions with $c^s_o>0$. %

\myparagraph{Using the feasibility scores during training for \settingNickNS. }
In \settingNickNS, we may obtain additional supervision by estimating missing labels. %
One straightforward way to achieve this is through %
pseudo-labeling \cite{lee2013pseudo}, a semi-supervised learning technique that takes the model predictions as ground-truth for unlabeled samples. In \settingNickNS, this means that, when the state (object) label is missing, pseudo-labeling will impute as label the object (state) predicted with the highest score. To avoid that the pseudo-labels form unfeasible compositions, we can use our prior to aid the pseudo-labeling process. %

Given either an object label $o$ or a state label $s$, we estimate their respective state and object pseudo-labels as: %
\begin{equation}
\label{eq:state-pseudo}
    \hat{s} \thicksim \mathtt{Gumbel}\left(\phi_s(z)\odot  c^o\right),\;\hat{o} \thicksim \mathtt{Gumbel}\left(\phi_o(z)\odot  c_s\right)
\end{equation}
where  %
$c^o$ is the vector containing the compatibility scores for all states given the object $o$, \ie $c^o = [c_s^o]_{s\in S}$, and $c_s$ is its counterpart for all objects given the state $s$, \ie $c_s = [c_s^o]_{o\in O}$\footnote{We assume $S$ and $O$ to be alphabetically ordered.}. %
Note that in both equations we %
sample the pseudo-labels using Gumbel-softmax ($\mathtt{Gumbel}$) \cite{jang2016categorical}. We found this choice helpful to make the model more robust to noisy predictions and less biased toward the training set latent label distribution.
Our objective function becomes:
\begin{align}
\begin{split}
\label{eq:loss-kvisprod}
    \mathcal{L}^{\text{\settingNickNS}}_{\text{visprod}} =& \sum_{(x_s,s)\in \mathcal{T}_s} \mathcal{L}_\text{state}(x_s,s) +\mathcal{L}_\text{obj}(x_s,\hat{o}) \\&+ \sum_{(x_o,o)\in \mathcal{T}_o} \mathcal{L}_\text{obj}(x_o,o) + \mathcal{L}_\text{state}(x_o,\hat{s}).
    \end{split}
\end{align}
We use this objective in 
\settingNick during training and Eq.\eqref{eq:loss-visprod} for OW-CZSL. In both cases we perform inference through Eq.\eqref{eq:inf-hm}. Since we couple independent primitive prediction with external knowlede to refine them, we name the method \fullnameBold\ (\nickMethodNS). Figure~\ref{fig:model-fig} illustrates our approach during training and inference.

{
\setlength{\tabcolsep}{6pt}
\renewcommand{\arraystretch}{1.1}
\begin{table*}[t]
    \centering
    \resizebox{0.9\linewidth}{!}{\begin{tabular}{c  l| c c c c | c c c c |  c c c c }
    \multicolumn{2}{c}{\multirow{2}{*}{\textbf{Method}}} & \multicolumn{4}{c}{\textbf{MIT-States}}& \multicolumn{4}{c}{\textbf{UT Zappos}}& \multicolumn{4}{c}{\textbf{C-GQA}}\\
                                                &   
                                                &S   & U    & HM    & AUC   
                                                &S   & U   & HM    & AUC 
                                                &S   & U    & HM    & AUC \\\hline
      &TMN~\cite{tmn}
      &  12.6        &  0.9     &   1.2 &   0.1  
     &     55.9  &18.1           & 21.7       &  8.4
     & NA  &    NA   & NA   & NA\\
    &AoP~\cite{aopp}     
     &16.6   &5.7       &4.7    &0.7 
     &  50.9 &  34.2     & 29.4   & 13.7
     & NA &  NA     &    NA&NA\\
      &LE+~\cite{redwine}  
      &14.2   &2.5       &2.7   &0.3     
     & {60.4}  &   36.5    & 30.5  & 16.3 
     & 19.2 & 0.7 & 1.0   &0.08\\
     & VisProd~\cite{redwine} & 20.9 & 5.8 & 5.6 &  0.7 & 54.6  & 42.8 & 36.9 & 19.7  & 24.8  & 1.7  & 2.8  & 0.33  \\ 
      &SymNet~\cite{symnet} 
      &21.4   &7.0       &5.8   &0.8     
     &53.3   &44.6       &34.5   &18.5
     &  26.7     &  2.2      & 3.3    &0.43\\
      &CompCos$^\text{CW}$~\cite{compcos}      
      &25.3   &5.5       &5.9   &0.9 
    &59.8   &45.6   &36.3   &20.8
    & 28.0 &    1.0   &    1.6&0.20\\
     &CGE$_\text{ff}$~\cite{cge} 
     & 29.6 & 4.0 & 4.9 & 0.7  
    & 58.8 & 46.5 & 38.0 & 21.5  
     &  28.3& 1.3      &    2.2&0.30\\
    &CompCos~\cite{compcos} 
     & {25.4}   &\textbf{10.0}       &\textbf{8.9}   &\textbf{1.6} & 
     59.3 & {46.8} & {36.9} & {21.3}
     &28.4  &      1.8 &    2.8&0.39\\
     &CGE~\cite{cge} 
     &  \textbf{32.4}& 5.1      &   6.0 & 1.0
     & 61.7      & 47.7  & 39.0      &  23.1  
      & \textbf{32.7} &   1.8    &   2.9 & 0.47\\
\hline
     & \nickMethodNS\textsubscript{ff} & 23.4 & 7.0 & 6.7 & 1.0 & 58.0  & 47.2  & 39.1  &  22.9 & 26.6  & 2.1  & 3.4  & 0.44  \\ 
     & \nickMethod & 28.4 & 7.5 & 7.4 &  1.3 & \textbf{61.8} & \textbf{52.1} & \textbf{42.3} & \textbf{26.5} & 31.5 & \textbf{2.9} & \textbf{4.7} & \textbf{0.78} \\ 
  \hline
    \end{tabular}}
    \vspace{1pt}
    \caption{\textbf{Open World CZSL results} on MIT-States, UT Zappos and C-GQA. We measure best seen (S) and unseen accuracy (U), best harmonic mean (HM), and area under the curve (AUC) on the compositions. {\nickMethodNS\textsubscript{ff} refers to our proposed method with a frozen backbone.} %
    }
    \vspace{-10pt}
    \label{tab:sota-open}
\end{table*}
}
\section{Experiments}
\vspace{-5pt}
\label{sec:exps}

\myparagraph{Datasets.} We use the three standard datasets for Compositional Zero-Shot Learning, namely UT-Zappos~\cite{utzappos1,utzappos2}, MIT-States~\cite{mitstates} and the recently proposed C-GQA~\cite{cge} dataset. \textbf{UT-Zappos} contains 12 object categories (shoe types) and 16 state categories (material types), with $83$ seen compositions and a compositonal space of 192 compositons. %
\textbf{MIT States} is a larger dataset that contains 245 object categories in 115 possible states. In total, it contains $1,262$ seen compositions and an output space of $28,175$ compositions in OW-CZSL. Finally, \textbf{C-GQA} is a recently proposed dataset\footnote{We refer to the updated split in \url{https://github.com/ExplainableML/czsl}.} with 674 object categories and 413 state categories. It contains $5,592$ training compositions and a full compositional space with $278,362$ compositions.

\myparagraph{Baselines.} %
In OW-CZSL, we compare \nickMethod against standard CZSL approaches; namely Attributes as Operators (AoP)~\cite{aopp}, Label Embed+ (LE+)~\cite{redwine}, Task Modular Networks (TMN)~\cite{tmn}, SymNet~\cite{symnet}, Compositional Graph Embeddings (CGE)~\cite{cge} and Compositional Cosine Logits (CompCos)~\cite{compcos}. In the tables, we refer to the closed world version of CompCos as CompCos$^\text{CW}$ and the variant of CGE with a frozen feature extractor as CGE\textsubscript{ff}. 

In \settingNickNS, we compare \nickMethod with CGE~\cite{cge} and CompCos~\cite{compcos}, the  state-of-the-art models in the closed and open-world settings respectively. These methods are adapted to \settingNick by marginalizing their predictions over states/objects, when the state/object information is available, minimizing the cross-entropy loss on the ground-truth annotation. %
We also experiment with popular semi-supervised learning techniques such as entropy minimization~\cite{grandvalet2005semi} and pseudo-labeling~\cite{lee2013pseudo}, adding them to both CompCos and CGE.

\myparagraph{Evaluation Protocol.}
For the OW-CZSL setting, we follow the standard splits of \cite{tmn,cge}, evaluate all the methods on the generalized  %
setting, where the model recognizes samples from both seen and unseen compositions. Following the protocol in \cite{tmn}, we apply a bias on the seen compositions at test time, measuring the performance as best seen~(S) and best unseen~(U) accuracy, best harmonic mean~(HM) as well as the area under the curve~(AUC) by varying the bias.%

 For the \settingNick setting, we propose a new split of the training split set, separating samples with object and state labels. This is done by keeping only the object label for half the samples, while for the remaining half, only the state label, ensuring that every object and state is seen in the training set.%
 Furthermore, for \settingNick setting, the model has no access to seen compositons $Y_s$. %
Thus, we evaluate the model in the full compositional space, without %
subtracting any bias on $Y_s$. %
Therefore, we use as metric the seen (S) and unseen (U) accuracy, and their harmonic mean (HM), as it is standard in Zero-Shot Learning~\cite{zslxian18benchmark}. 

\myparagraph{Implementation Details.}
We follow the standard practices in the CZSL literature~\cite{cge,compcos}, by using a ResNet18~\cite{resnet} feature extractor. For the state and object classifiers, we {follow \cite{cge} and} use Multi-Layer Perceptrons with three layers, comprising Layer Normalization~\cite{ba2016layer} and Dropout~\cite{srivastava2014dropout}. The model is optimized using Adam~\cite{kingma2014adam} with the default hyperparameters, %
a learning rate and weight-decay of $5e$-$5$. %

\subsection{Open-World CZSL}
\vspace{-5pt}
The results on the challenging OW-CZSL setting are reported in Table~\ref{tab:sota-open}. %
In this setting \nickMethod either outperforms or it is competitive with the state of the art. Specifically, on UT-Zappos {\nickMethodNS} outperforms the best competitor (CGE) in all metrics, with an improvement of 3.4\% in AUC (26.5 vs 23.1), 3.3\% in best HM (42.3 vs 39.0) and 4.4\% best unseen (52.1 vs 47.7). Similarly, without end-to-end training, \nickMethodNS$_\text{ff}$ surpasses the best baseline (CGE$_\text{ff}$) by 1.3 in AUC (22.9 vs 21.5) and by 1.3 in best HM (39.1 vs 38.0). %

These results are confirmed in the challenging C-GQA dataset. Despite an output space of almost $280$k compositions, \nickMethod obtains 0.78 AUC vs 0.47 of the best competitor (CGE), 4.7 HM (vs 3.3 of SymNet) and 2.9 on best unseen (vs 2.2 of SymNet). When non-finetuned, the method achieves competitive results \wrt SymNet (\eg 3.4 HM) while being much easier to optimize, since \nickMethod does not impose any constraint on the compositional space. Our results indicate that modeling states and objects independently may be an effective approach to deal with the very large output space of OW-CZSL. This independence assumption ensures that each predictor learns a discriminative classifier over a few hundred classes rather than a single classifier over thousands of compositions, which does not scale well even with powerful architectures (\eg graph-convolutional neural network of \cite{cge}) and initialization through side-information (\eg word-embeddings \cite{word2vec}).

Finally, the table also highlights %
the gap between \nickMethod and the VisProd baseline of \cite{redwine}. In particular, our revised model (without fine-tuning) consistently surpasses VisProd in AUC (\ie 1.0 vs 0.7 on MIT-States, 22.8 vs 19.7 in Zappos, 0.44 vs 0.33 in C-GQA) and best harmonic mean (6.7 vs 5.6 on MIT-States, 39.3 vs 36.9 on Zappos and 3.4 vs 2.8 on C-GQA). These results confirm the importance of our modifications to the original VisProd model, as we will ablate in the following.

\setlength{\tabcolsep}{3pt}
\renewcommand{\arraystretch}{1.2}
\begin{table}[t]
    \centering
    \begin{tabular}{l| c | c c c c }
       & Marginaliz.   &Seen   & Unseen    & HM & AUC    \\\hline
        \multirow{2}{*}{CGE\textsubscript{ff}} && \textbf{25.5}& 5.7&6.5 & 1.0 \\
        &\cmark& 24.0 & \textbf{7.8}& \textbf{8.1} & \textbf{1.3}\\\hline
                  
        \multirow{2}{*}{CGE}  && \textbf{27.2}  & 6.6 & 7.0 & 1.3 \\
         &\cmark& 25.1& \textbf{8.1} & \textbf{8.1} & \textbf{1.4}\\\hline
                  
    \end{tabular}
    \vspace{1pt}
    \caption{OW-CZSL results in the %
    validation set of MIT-States when using marginalization. {CGE\textsubscript{ff} is the approach of \cite{cge} with frozen backbone whereas CGE performs end-to-end training.} }%
    \vspace{-8pt}
    \label{tab:marginalization}
\end{table}

 \begin{figure}[t]
 \centering
 \centering
 \includegraphics[width=0.9\linewidth]{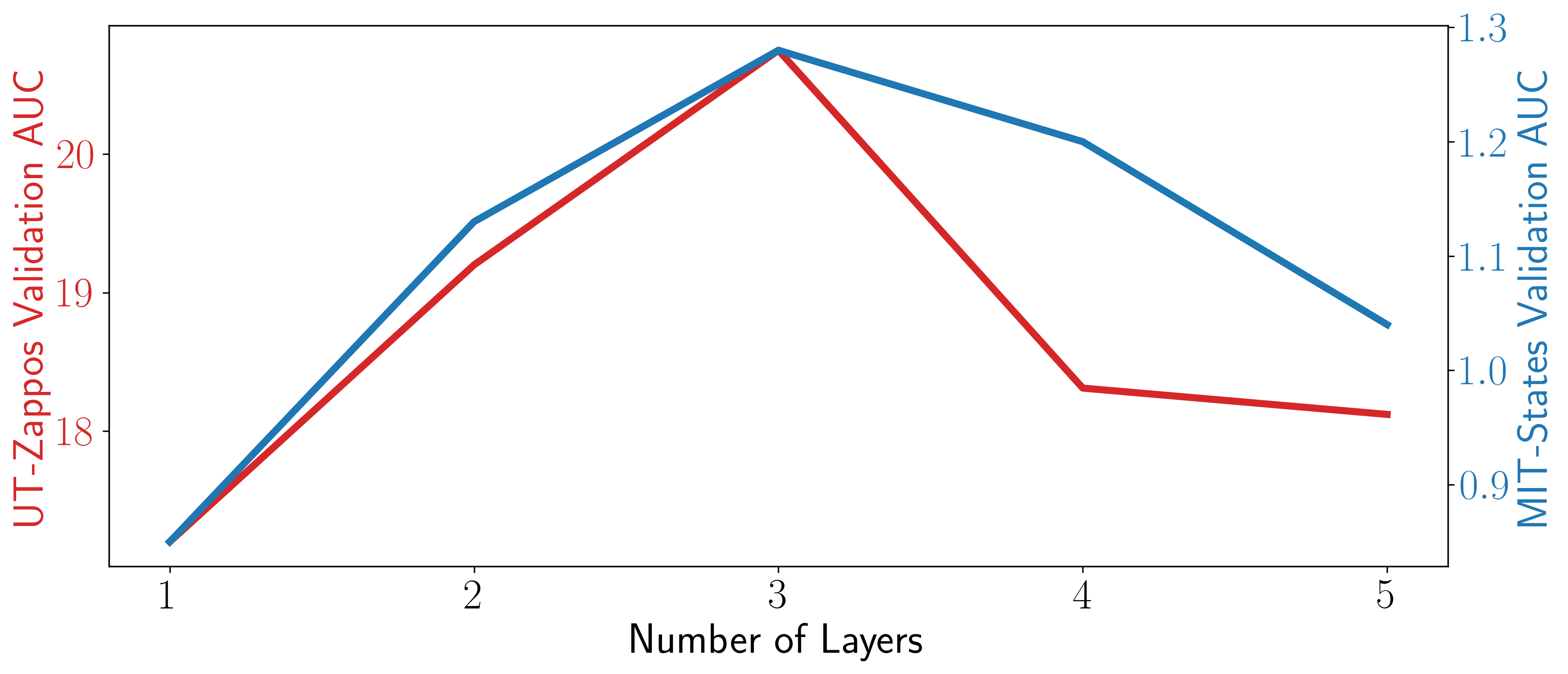}
 \label{fig:top1_1}

\caption{Ablation study on the importance of the depth of the object and state-classifiers for \nickMethodNS$_\text{ff}$
on UT-Zappos (red curve) and MIT-States (blue curve) validation set for the OW-CZSL setting. Performance is measured in AUC.}
 \label{fig:depthplot}
 \vspace{-8pt}
 \end{figure}

\setlength{\tabcolsep}{3pt}
\renewcommand{\arraystretch}{1.2}
\begin{table}[t]
    \centering
    \begin{tabular}{l| c | c c c c }
       & Mask   &Seen   & Unseen    & HM & AUC    \\\hline

        \multirow{2}{*}{VisProd}  && \textbf{24.8} & 6.8 & 7.3 & 1.1 \\
         &\cmark& 24.7 & \textbf{7.2} & \textbf{7.6} & \textbf{1.2} \\\hline                 
                  
         \multirow{2}{*}{\nickMethodNS\textsubscript{ff}}  && 26.3& 7.4 & 7.9 & 1.3\\
          &  \cmark & \textbf{26.5} & \textbf{7.7} & \textbf{8.2} & \textbf{1.4}  \\\hline
    \end{tabular}
    \vspace{1pt}
    \caption{OW-CZSL results on MIT states validation set while applying our feasibility-based masks ($f_{\text{KG}}$) on different models.}
    \vspace{-15pt}
    \label{tab:hardmask-open}
\end{table}

\vspace{-5pt}
\subsubsection{Why \nickMethod works in OW-CZSL?}
\textbf{Separately predicting objects and states.} We argue that an important reason for the success of \nickMethod is the separate treatment of states and objects, As stated in previous discussion, predicting state and objects independently makes the OW-CZSL problem easier and more scalable \wrt predicting directly over thousands of compositions. To verify this hypothesis, we take the state-of-the-art method in standard CZSL, CGE \cite{cge} %
(with and without end-to-end training) 
and we modify its classifier in such a way that it can output separate objects and states. Specifically, we take the state predictions by marginalizing their scores across all possible objects and, similarly, we marginalize object predictions over the set of possible states.  %
Results are reported in Table~\ref{tab:marginalization}. We see a consistent increase in best unseen accuracy (5.7 vs 7.8 for CGE\textsubscript{ff}, 6.6 vs 8.1 for CGE) and in the best HM (6.5 vs 8.1 for CGE\textsubscript{ff}, 7.0 vs 8.1 for CGE) when we separate the two predictions. As a consequence, the methods also improve in AUC (1.1 to 1.2 CGE\textsubscript{ff}, 1.3 to 1.4 for CGE). %
This indicates how providing a separate ground for objects and states predictions is %
a useful strategy in the open-world setting. Operating in the primitives rather than the compositional space, provides a simplification of the problem that can improve the performance even of existing state-of-the-art CZSL models.

{
\setlength{\tabcolsep}{6pt}
\renewcommand{\arraystretch}{1.1}
\begin{table*}[t]
    \centering
    {\begin{tabular}{l  l| c c c  | c c c  |  c c c  }
    \multicolumn{2}{c}{\multirow{2}{*}{\textbf{Method}}} & \multicolumn{3}{c}{\textbf{MIT-States}}& \multicolumn{3}{c}{\textbf{UT Zappos}}& \multicolumn{3}{c}{\textbf{C-GQA}}\\
                                                &   
                                                &S   & U    & HM      
                                                &S   & U   & HM     
                                                &S   & U    & HM    \\\hline
   {CGE$_\text{ff}$~\cite{cge}} &
     & 19.6 & 1.3 & 2.4    
    & 50.3 & 3.4 & 5.0 
     &  17.4 & 0.4 & 0.9 \\
    &+Pseudo-Lab.
     & \textbf{19.7}   & 0.9     & 1.8    
     & 48.5   & 1.1 & 2.2 
     & 19.8   & 0.2 & 0.4\\
    &+Entropy Min.
     & 15.1   & 1.7     & 3.1    
     & 51.9   & 4.2 & 6.4 
     & 22.1   & 0.4 & 0.9\\\hline
   {CompCos~\cite{compcos}}& 
     & 10.8   & 2.0     & 3.6    
     & 52.4   & 4.1 & 7.6 
     & 24.3   & 0.4 & 0.7\\
    &+Pseudo-Lab.
     & 9.2   & 1.9     & 3.2    
     & 52.9   & 3.7 & 6.8 
     & 23.6   & 0.3 & 0.5\\
    &+Entropy Min.
     & 13.2   & 2.1     & 3.7    
     & 55.0   & 4.2 & 7.9 
     & 23.1   & 0.6 & 1.1\\\hline
     {CGE~\cite{cge}}&
     &  17.9 & 1.6     &    3.0
     & 55.8      & 5.9  & 10.7        
      & 25.6 &   0.7    &   1.4\\ 
    &+Pseudo-Lab.& 10.6 & 2.3    & 3.8   
     &  56.1     & 3.9  & 7.3      
      & 21.3 &      0.6 & 1.2 
     \\ 
       &+Entropy Min.& 17.8 &1.6     &3.0    
     &  60.1     & 4.7  & 8.7        
      & 24.8 & 1.0      &  1.8
     \\\hline

     \multicolumn{2}{c|}{\hspace{+5pt}\nickMethodNS\textsubscript{ff}} & 13.5 & \textbf{2.6} & \textbf{4.4}  & 53.8  & 6.9  & 12.3   & 22.3  & 0.9  & 1.7    \\ 
     \multicolumn{2}{c|}{\nickMethodNS}  & 18.4 & 2.2 & 4.0 &   \textbf{57.9} & \textbf{7.4} & \textbf{13.1}  & \textbf{26.9} & \textbf{1.2} & \textbf{2.3}  \\ 
  \hline
    \end{tabular}}
    \vspace{2pt}
    \caption{\textbf{Partial Open-World CZSL results} on MIT-States, UT Zappos and C-GQA. We measure seen (S) and unseen accuracy (U) on the compositions and their harmonic mean (HM). \nickMethod refers to our full model with our knowledge-guided pseudo-labeling and inference. CGE$_\text{ff}$ and \nickMethodNS$_\text{ff}$ denotes the non-finetuned version of the methods. For each CZSL baseline, we show the results of the original methods and the same coupled with Entropy-Minimization (Entropy Min.)~\cite{grandvalet2005semi} or Pseudo-labeling (Pseudo-Lab.)~\cite{lee2013pseudo}. %
    }
    \vspace{-15pt}
    \label{tab:sota-partial}
\end{table*}
}
\begin{table}[t]
    \centering
\begin{tabular}{@{}l|cccc@{}}
Method                      & Seen & Unseen & HM   \\\hline
\nickMethodNS                & \textbf{16.6}  &  2.8  & 4.8     \\
  \hspace{+10pt}+ Pseudo-Labeling             & 15.9 & 2.7   & 4.6      \\
    \hspace{+20pt}+ Gumbel Softmax        &  16.1 &  2.6  & 4.5     \\
      \hspace{+30pt}+ ConceptNet-scores & \textbf{16.6} & \textbf{3.1}    & \textbf{5.3}  \\ \hline
\end{tabular}
\vspace{2pt}
\caption{\textbf{Partial Open-World CZSL results} on MIT-States validation set for different methods %
in terms of Seen, and Unseen accuracy and their Harmonic Mean (HM). Pseudo-Labeling, sampling the pseudo-label using Gumbel-Softmax, and using ConceptNet to filter unfeasible labels are added one after another.} %
\label{tab:gumbel-ablation}
\vspace{-15pt}
\end{table}

\vspace{1pt}
\myparagraph{Effect of depth of the classifier on \nickMethodNS.} We ablate the impact of the depth in Fig~\ref{fig:depthplot} for {\nickMethodNS\textsubscript{ff}} on both MIT-States and UT-Zappos validation sets. The validation AUC on both UT-Zappos (red curve) and MIT-States (blue curve) rapidly increases with the depth of the classifier. This shows the importance of taking into account the multi-task nature of the problem and instantiate objects and state classifiers that have enough capacity to extract primitive-specific features. %
While deeper predictors help, after 3 layers the performance degrades (\ie, going from 26.9 to 24.3 on UT-Zappos when depth is increased from 3 to 5 layers). The reason behind this drop is mainly linked to overfitting on seen compositions.

\vspace{1pt}
\myparagraph{Effect of the knowledge-based masking.} We ablate the impact of masking out unfeasible compositions from the output space in Table \ref{tab:hardmask-open} for MIT-states validation set. We experiment with both VisProd and \nickMethodNS\textsubscript{ff}. As the table shows, \nickMethodNS\textsubscript{ff} benefits from removing unfeasible compositions, with the unseen accuracy going from 7.3 to 7.5 and the best HM from 7.6 to 8.1. Similarly, %
also for VisProd we see a consistent improvement for both the unseen accuracy (6.8 to 7.2) and the best HM (7.3 to 7.6) when its output space is filtered.
These results show how removing unfeasible compositions from the output space benefit the performance of OW-CZSL models and that ConceptNet is a reliable source for estimating feasibility scores of the compositions. %

\subsection{CZSL under Partial supervision}
The results on our proposed \settingNick setting are reported in Table~\ref{tab:sota-partial}. In addition to recognizing unseen compositions in an extremely large compositional space, in \settingNick the model has to cope with the lack of %
compositional labels. 

In this scenario \nickMethod achieves state-of-the-art results on all the three datasets. %
On UT-Zappos, \nickMethod  achieves a HM of 13.1 vs 10.7 of the best competitor (CGE). %
Similarly, on MIT-States and C-GQA, the results obtained by \nickMethod exceed the performance of the competitors, with \nickMethod achieving 4.4 HM vs 3.7 of CompCos with entropy minimization on MIT-States, and 2.3 (1.7 when non end-to-end trained) HM on C-GQA vs 1.8 of the best competitor (CGE). %
It is interesting to note %
how, even incorporating semi-supervised learning techniques in CGE and CompCos does not %
bridge the gap between \nickMethod and these methods. Interestingly, semi-supervised learning techniques do not bring consistent improvements across CZSL models and tasks. %
{Entropy minimization, by pushing the method %
toward confident predictions, improves CGE (1.8 vs 1.4 AUC) and CompCos (1.1 vs 0.7) on C-GQA, and improves CGE$_\text{ff}$ on MIT-States (3.1 vs 2.4 AUC). However, in the other settings either achieves the same performances of the baseline (\eg CGE on MIT-States) or degrades them (\eg CGE on UT-Zappos, 8.7 vs 10.7 AUC). On the other hand, pseudo-labeling degrades the performance of the CZSL models in all settings (\eg from 3.6 to 3.2 AUC of CompCos on MIT-States), providing advantages only for CGE on MIT-States (3.8 vs 3.0 AUC). %
The low efficacy of standard semi-supervised learning techniques, is due to the fact that the top predicted object/state may form an unfeasible composition with the ground-truth state/object. Without modeling the feasibility of state-object compositions, pseudo-labeling may assign incorrect (and unfeasible) labels while entropy minimization may push the model toward incorrect predictions.} %
In the following section, we discuss why this happens and why it is important to restore to our knowledge-aided pseudo-labeling in \settingNickNS.

\vspace{-10pt}
\subsubsection{Ablation Study}

We ablate the two important components of our proposed approach: the pseudo-labeling strategy and the quality of the ConceptNet feasibility scores. 

\myparagraph{Effect of the pseudo-labeling strategy.} In Table~\ref{tab:gumbel-ablation}, we consider a few alternatives to our proposed pseudo-labeling strategy for \nickMethodNS. %
The first option is to just introduce vanilla pseudo-labeling. This is done by replacing the missing labels with the top predicted class, when %
the confidence of the model is greater than a threshold. This strategy has a negative effect, reducing the accuracy of the model on both seen and unseen compositions (\ie 16.6 vs 15.9 on the seen, 2.8 vs 2.7 on the unseen). {This is because pseudo-labeling alone is prone to confirmation bias \cite{arazo2020pseudo} since the model can become increasingly more confident about the predictions that, without modeling the feasibility of compositions, may be not only incorrect, but also unfeasible.}%

The second alternative is to use Gumbel-softmax \cite{jang2016categorical}, sampling a label instead of performing direct pseudo-labeling. This approach may allow the model to overcome the confirmation bias, especially in the initial training stages by sampling different labels \wrt the top prediction. However,  %
this strategy alone does not achieve good results, degrading the performance of the base model (\eg 4.5 vs 4.8 HM) and performing slightly worse than standard pseudo-labeling (4.5 vs 4.6 HM). %
This performance degradation is due %
in both cases by not exploiting the external knowledge to correct the pseudo-labeling process, avoiding using incompatible compositions as supervision. %

Our pseudo-labeling strategy, on %
the other hand, provides a clear improvement (5.3 vs 4.8 HM) over our method %
which already attains state-of-the-art results on open-world CZSL benchmarks. {Its benefits come from using external knowledge to assign a feasibility score to each composition. These feasibility scores are used to refine the pseudo-labeling process, avoiding incompatible compositional labels. In addition to this, using the model scores to sample the label in a probabilistic manner ensures that confirmation bias and model drift can be avoided. }%

\setlength{\tabcolsep}{2pt}
\renewcommand{\arraystretch}{1.2}
\begin{table}[t]
    \centering
    \resizebox{\linewidth}{!}{
    \begin{tabular}{ l  l@{\hskip 0.12in}  l }
    \hline
        & \multicolumn{2}{c}{\textbf{Compositions}} \\
       & Most Feasible (Top-5)  &  Least Feasible (Bottom-5)   \\\hline
       & painted paint & grimy balloon  \\
        &muddy mud & steaming bracelet \\
         & frozen ice & blunt clock \\
          &mossy moss &  thin garage \\
           &cloudy cloud&  unpainted belt  \\
           \hline
        \textbf{Objects} & \multicolumn{2}{c}{\textbf{States}} \\
        & Most Feasible (Top-3)  &  Least Feasible (Bottom-3) \\
        \hline
         chains & frayed, broken, loose & pureed, unripe, steaming\\
        sugar &melted, whipped, caramelized& scratched, ancient, coiled\\
         sword & blunt, sharp, splintered  & filled, runny, closed\\
         laptop&small, shattered, modern & cloudy, sunny, dull\\
         tulip&bright, wilted, ruffled& grimy, raw, damp\\ \hline 
    \end{tabular}
    }
    \vspace{1pt}
    \caption{Examples of ConceptNet feasibility scores. Top: Top-5 (left) and Least-5 (right) compositions per feasibility; %
    Bottom: Top-3 highest and Bottom-3 lowest feasible states per random objects.}
    \vspace{-12pt}
    \label{tab:most-feasible-general}
\end{table}

\myparagraph{Quality of the ConceptNet feasibility scores.} An important aspect of \nickMethod is the quality of the estimated feasibility scores.  %
In Table~\ref{tab:most-feasible-general} we show some qualitative results of our strategy, reporting detailed quantitative analyses on the supplementary material. ConceptNet assigns the highest feasibility scores to compositions where the state and the objects share the same root, such as \textit{painted paint}, \textit{muddy mud}, and \textit{mossy moss}. This is a consequence of the very similar contexts in which such states and objects appear. On the other hand, among the least feasible compositions we find objects with incompatible physical transformations (\ie \textit{steaming bracelet}), or unclear states (\eg \textit{unpainted belt}, \textit{blunt clock}). These scores reflect the reliance of ConceptNet relatedness scores to the context in which words appear. This is also a limitation, since rare co-occurrences can be deemed as unfeasible compositions (\eg \textit{grimy balloon}, \textit{thin garage}).%

When inspecting the most and least feasible compositions for random objects (Table~\ref{tab:most-feasible-general}, bottom), we find that the compositions ranked as most feasible such as \textit{bright tulip}, \textit{caramelized sugar}, and \textit{blunt sword} are indeed feasible in the reality, while the compositions marked as least feasible are not. Interestingly, the unfeasible compositions merge objects and states from different categories. Examples are food vs tools (\eg \textit{pureed chains}, \textit{scratched sugar}) and weather vs objects (\eg \textit{cloudy laptop}). This suggests that the ConcepNet-based feasibility scores encode a high-level notion of grouping, where objects and states can be separated depending on the contexts where they commonly occur. %

\vspace{-2pt}
\section{Conclusion}
\vspace{-3pt}
In this work, we addressed the problem of Open-World Compositional Zero-Shot learning (OW-CZSL), where the goal is to recognize compositions of objects and states in images given only a subset of them %
during training and without any prior on unseen compositions at test time. %
We address the problem by revisiting a simple CZSL method, %
VisProd, that independently predicts state and object labels. The idea is to simplify the problem by exploiting the much smaller cardinality of object and state sets \wrt the compositional labels. %
Our model, \nickMethodNS, uses two different feature extractors to account %
for the dissimilarity between the two tasks, %
and uses external information to remove less feasible compositions from the output space at test time. As \nickMethod does not require compositional labels, we explore a new challenging setting, CZSL under partial supervision (\settingNickNS) where training images have either only object or state annotation. In \settingNickNS, we use the feasibility scores to aid the estimation of the missing labels. Experiments show that \nickMethod achieves the state of the art in OW-CZSL and \settingNickNS, outperforming recent CZSL approaches coupled with standard semi-supervised learning techniques. %

\myparagraph{Limitations and Broader Impact.}
One weakness of our approach, shared with %
CZSL methods, %
is that the absolute performance on all OW-CZSL benchmarks is quite low (\eg 2.9 unseen accuracy on C-GQA). This can significantly hamper the real-world deployment of these models. However, we believe that this research topic is crucial to bridge the gap between machine and human visual compositional recognition abilities: our work contributes to the field by making another step toward this direction. %
Another limitation of our work is that automatic retrieving %
feasible state-object compositions is a process without expert supervision, {thus vulnerable to inaccuracies in the knowledge base. For instance, if a valid composition
is marked as unfeasible, it will not be predicted during inference and removed from candidate pseudo-labels during
pCZSL training}. %
This may lead to the model {producing erroneous outcomes, even %
reflecting potential biases in the knowledge base}. %
{Modeling issues in the knowledge base and/or merging multiple external sources} is an important topic for future research in OW-CZSL and pCZSL.

\vspace{5pt}
\myparagraph{Acknowledgments}
This work has been partially funded by the ERC (853489 - DEXIM), by the DFG (2064/1 – Project number 390727645), and as part of the Excellence Strategy of the German Federal and State Governments. %
The authors thank the International Max Planck Research School for Intelligent Systems (IMPRS-IS) for supporting Shyamgopal Karthik. 

{\small
\bibliographystyle{ieee_fullname}
\bibliography{egbib}
}
\appendix

{\section{Further implementation details}}
We set all hyperparameters of KG-SP following previous works \cite{compcos,cge} and our MLPs from the implementation of \cite{cge}, using 3 layers with dimension 768 in the first,
1024 in the second, the number of objects/states in the last, and intermediate Dropout layers with ratio 0.5. For consistency with previous works (\eg \cite{aopp,symnet,cge}), we also test our model without fine-tuning the backbone, denoting it as KG-SP\textsubscript{ff} in the tables.

\vspace{5pt}
\section{Analysis of the ConceptNet Embeddings}

In addition to the analysis presented in the main paper, we further analyze the quality of the ConceptNet embeddings both quantitatively and qualitatively. 

\myparagraph{Comparison with Alternative Word Embeddings.}
\label{sec:concept-thresh}
We compare ConceptNet~\cite{conceptnet} with other popular word embeddings such as Word2Vec~\cite{word2vec}, Glove~\cite{pennington2014glove} and FastText~\cite{fasttext}, as well as language models \cite{gpt2}. 

For what concerns the word embeddings, we reject compositions whose cosine similarity between the state and object embeddings is less than 0. For the language models, we use GPT-2 \cite{gpt2} by querying a specific prompt and extracting the likelihood of the next word to be either \texttt{Yes} or \texttt{No}. We can then set as feasible all compositions whose likelihood of \texttt{Yes} is higher than the likelihood of \texttt{No}. We tested several prompts (\eg \texttt{Can OBJ be STA?}, \texttt{You can see a STA OBJ}), but we found the most effective to be: \texttt{Question: Is this a STA OBJ? Answer:}, with \texttt{STA} and \texttt{OBJ} being a state and object respectively. %

\begin{figure*}[t]
\centering
\begin{subfigure}{0.49\linewidth}
\centering
\includegraphics[width=\linewidth]{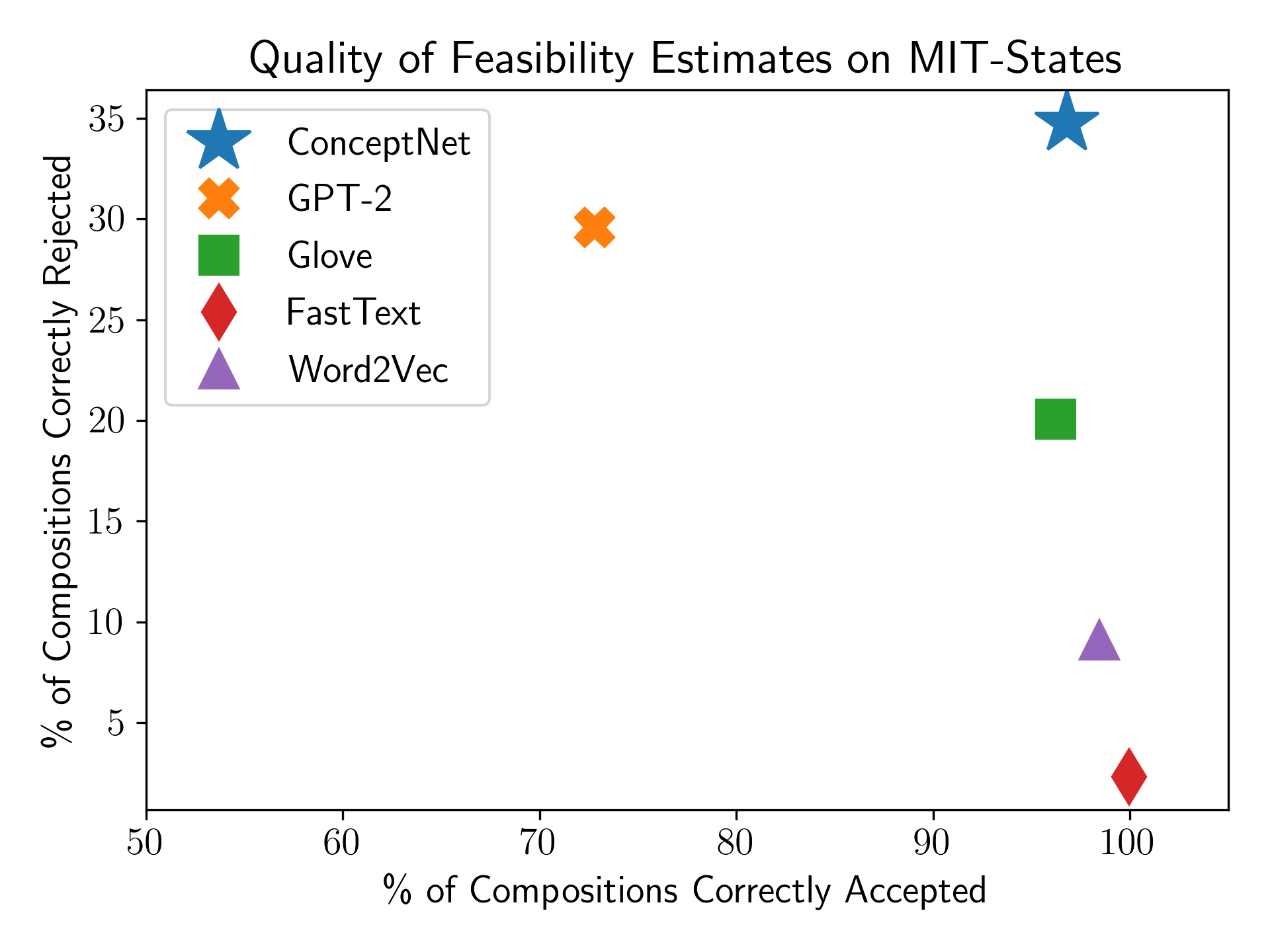}
\caption{MIT-States}
\end{subfigure}
\begin{subfigure}{0.49\linewidth}
\centering
\includegraphics[width=\linewidth]{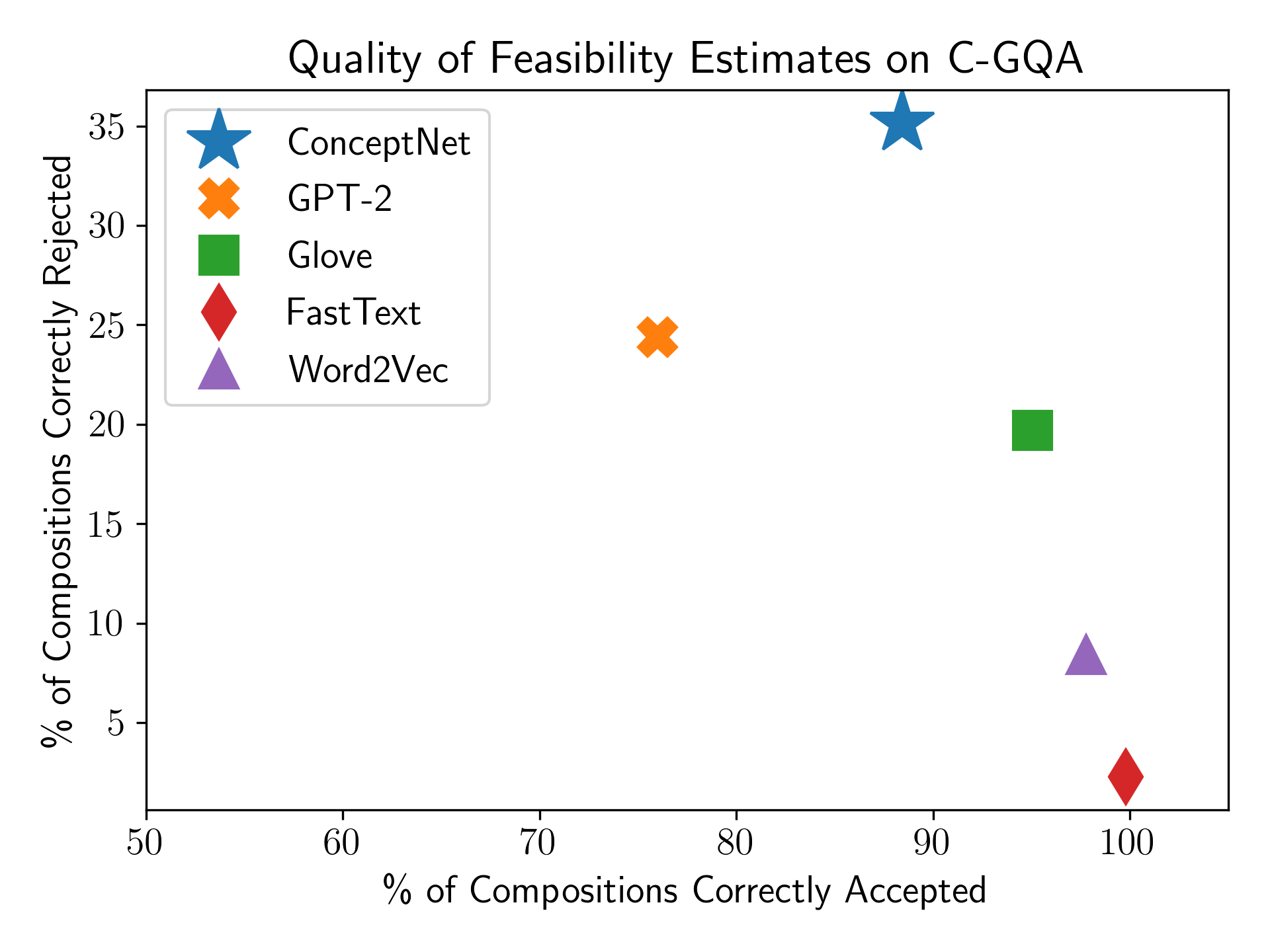}
\caption{C-GQA}
\end{subfigure}

\caption{{Comparison of the ConceptNet embeddings against other word embeddings (Word2Vec, Glove, FastText) and GPT-2. The x-axis denotes the percentage of compositions in the dataset that are correctly accepted. The y-axis denotes the percentage of compositions not present in the dataset that are correctly rejected. We can see that ConceptNet provides an excellent trade-off by rejecting a large number of unfeasible compositions, while still accepting most of the compositions present in the dataset.}}%
\label{fig:conceptnet-feas}
\end{figure*}

In Fig.~\ref{fig:conceptnet-feas}, we report the results of the feasibility scores computed through the various strategy. Specifically, we report in the x-axis the percentage of compositions correctly considered as feasible (w.r.t. the ones present in the dataset) and on the y-axis the percentage of compositions correctly rejected (\ie compositions not present in the datasets). We report these results for both MIT-states (Fig. 1.a) and C-GQA (Fig. 1.b). %
In both these datasets, FastText accepts nearly all the compositions therefore, rejecting very few compostions. Similar trends can be seen for Word2Vec and Glove, which reject more unfeasible compositions. While GPT-2 rejects more unfeasible compositions, it also rejects compositions present in the dataset. ConceptNet achieves the best trade-off by accepting most of the compositions present in the dataset, while still rejecting the largest possible number of unfeasible compositions among the competitors.

\begin{table}[]
\centering
\begin{tabular}{l|cccc}
Feasibility scores              & Seen & Unseen & HM  & AUC \\ \hline
None                & 26.3 & 7.4    & 7.9 & 1.3 \\
CompCos (No Tuning) & 26.3 & 7.4    & 7.9 & 1.3 \\
CompCos (Tuned)     & 26.3 & 7.5    & 8.0 & \textbf{1.4} \\
ConceptNet          & \textbf{26.5} & \textbf{7.7}    & \textbf{8.2} & \textbf{1.4} \\ \hline
\end{tabular}
\vspace{2pt}
\caption{Comparison of various feasibility scores on the MIT-States for KG-SP\textsubscript{ff}. We see that despite tuning the threshold for CompCos, ConceptNet is still able to achieve better results.}
\label{tab:compcosfeas}
\end{table}

\myparagraph{Quantitative Analysis.}
In CompCos \cite{compcos}, a method for estimating the feasibility scores of each composition is presented. CompCos computes feasibility scores by using the cosine similarity of learned object/state embeddings, exploiting the available seen compositions during training. Specifically, for a given unseen state-object composition, CompCos computes the cosine similarity of the embeddings of the given object with those of any object containing the same state as seen during training, taking the maximum similarity as feasibility score. The same procedure is applied for the states, and the two scores are averaged.

A natural question is whether our ConceptNet scores can outperform the ones of CompCos. %
We analyze this in Table \ref{tab:compcosfeas} for OW-CZSL on the MIT-States validation, using different feasibility scores to perform hard-masking on KG-SP$_\text{ff}$, . From the table, we can see that %
hard masking with the feasibility scores of CompCos %
does not bring any benefits when used with a natural threshold of zero (\ie half of the range of cosine-similarity scores). When we tune a threshold to filter out less feasible compositions (as in \cite{compcos}), we see slight improvements (\eg 8.0 vs 7.9 AUC, 1.4 vs 1.3 best HM). However, our ConceptNet-based feasibility scores bring a larger improvement in performance (\eg 7.7 best unseen accuracy, 8.2 best HM), without requiring tuning any threshold or accessing compositional annotations during training. The latter feature makes ConceptNet feasibility scores applicable also in pCZSL, while CompCos-based ones cannot be applied in such setting.  %

\myparagraph{Qualitative Analysis.}
In this section, we  expand the analysis of Section 4.2.1 in the main paper, and we present the top-3 most feasible states and bottom-3 least feasible states for 25 randomly selected objects in Table~\ref{tab:most-feasible-general}. Similarly, we show the top-3 most feasible objects and the bottom-3 least feasible objects for 25 randomly selected states in Table~\ref{tab:supp-most-feasible-state}
. %
As discussed earlier both in Section 4.2.1 in the main paper and in Section~\ref{sec:concept-thresh}, ConceptNet provides good estimates of the feasibility of state-object compositions. %
Overall, the most feasible compositions selected by the embeddings usually turn out to be commonly occurring combinations of the states and objects. For instance, \textit{clear sky}, \textit{burnt flame}, and \textit{engraved jewelry} are all frequently occurring state-object combinations. An interesting aspect of the ConceptNet embeddings is the implicit clustering of the most relevant object/state pairs. For instance, in Table \ref{tab:supp-most-feasible-general} food items (\textit{paste}, \textit{pizza}, \textit{salad}) are linked to cooking-related states states such as \textit{sliced} and \textit{diced}. Similarly, both \textit{chain} and \textit{cord} are linked to the state \textit{frayed}. We can see similar patterns also in Table~\ref{tab:supp-most-feasible-state},  where \textit{murky} and \textit{muddy} are linked with \textit{mud}, and \textit{unripe} with \textit{fruit}, \textit{pear} and \textit{orange}.  %
 However, the estimated feasibility scores are not perfect, and can lead to erroneous outcomes. For instance, this happens for rarely occurring compositions that might be considered unfeasible %
 (\eg  \textit{steaming chains} in Tab.~\ref{tab:most-feasible-general}, \textit{dirty bracelet} in Tab.~\ref{tab:supp-most-feasible-state}), and for co-occurring words that might receive high feasibility scores despite not being compatible (\eg \textit{sunny sea} in Tab.~\ref{tab:supp-most-feasible-general}, \textit{molten flame} in Tab.~\ref{tab:supp-most-feasible-state}). %
 This is a limitation of feasibility scores based on single words representations, since these models are biased by the context in which word appear, and their co-occurring frequency. %
 Using a combination of cues from multiple sources may be an effective tool to deal with this issue. %

\setlength{\tabcolsep}{2pt}
\renewcommand{\arraystretch}{1.2}
\begin{table}[t]
    \centering
    \resizebox{\linewidth}{!}{
    \begin{tabular}{ l  l@{\hskip 0.12in}  l }
        \textbf{Objects} & \multicolumn{2}{c}{\textbf{States}} \\
        & Most Feasible (Top-3)  &  Least Feasible (Bottom-3) \\
        \hline
balloon & inflated, deflated, filled & grimy, rusty, raw \\
bear & heavy, large, huge & crinkled, dark, damp \\
bridge & narrow, curved, bent & barren, cluttered, pressed \\
building & standing, tall, moldy & ruffled, sharp, pureed \\
bush & mossy, blunt, wilted & ripped, broken, grimy \\
cable & coiled, frayed, loose & empty, filled, barren \\
camera & raw, sharp, lightweight & pierced, crinkled, ruffled \\
chains & loose, broken, frayed & pureed, unripe, steaming \\
chair & standing, upright, bent & pierced, thawed, sliced \\
concrete & unpainted, crushed, molten & ruffled, folded, whipped \\
cord & coiled, frayed, loose & foggy, dirty, grimy \\
flower & wilted, ruffled, verdant & grimy, cored, scratched \\
jewelry & engraved, pierced, worn & steaming, squished, full \\
mirror & painted, dented, shiny & unripe, pureed, cooked \\
paste & mashed, pureed, sliced & tall, standing, winding \\
pizza & sliced, cooked, diced & bright, clear, modern \\
salad & diced, mashed, sliced & tight, narrow, smooth \\
sea & sunny, murky, fresh & closed, unpainted, squished \\
shower & wet, damp, steaming & burnt, cored, thin \\
snake & coiled, winding, curved & creased, crumpled, pressed \\
steel & molten, rusty, shiny & runny, unripe, verdant \\
steps & standing, winding, straight & viscous, tight, dry \\
stream & muddy, winding, spilled & dented, rusty, caramelized \\
sugar & caramelized, whipped, melted & scratched, ancient, coiled \\
tile & painted, unpainted, moldy & whipped, inflated, ripe \\\hline

    \end{tabular}
    }
    \vspace{1pt}
    \caption{Examples of top-3 and bottom-3 states associated to 25 randomly selected objects, according to ConceptNet feasibility scores.}
    \label{tab:supp-most-feasible-general}
\end{table}

\setlength{\tabcolsep}{2pt}
\renewcommand{\arraystretch}{1.2}
\begin{table}[t]
    \centering
    \resizebox{0.935\linewidth}{!}{
    \begin{tabular}{ l  l@{\hskip 0.12in}  l }
        \textbf{States} & \multicolumn{2}{c}{\textbf{Objects}} \\
        & Most Feasible (Top-3)  &  Least Feasible (Bottom-3) \\
        \hline
ancient & stone, bronze, ceramic & candy, sandwich, cake \\
bright & lightbulb, sky, orange & sandwich, pizza, beef \\
browned & sauce, butter, beef & key, vacuum, wall \\
brushed & coat, wool, dust & cave, lake, building \\
burnt & flame, fire, smoke & bathroom, shower, camera \\
clear & sky, concrete, glass & sandwich, pants, pizza \\
cooked & meat, soup, sauce & key, mirror, moss \\
crushed & ground, lemon, concrete & lake, pond, clock \\
curved & blade, steel, knife & oil, eggs, bag \\
deflated & balloon, bubble, tire & kitchen, wood, coffee \\
dirty & dirt, mud, dust & balloon, bracelet, cord \\
dull & blade, knife, sword & fig, mountain, cookie \\
filled & vacuum, bag, bucket & cable, cliff, sword \\
heavy & metal, bear, handle & fig, pool, pond \\
moldy & basement, dust, carpet & tiger, road, highway \\
molten & metal, flame, copper & book, cat, furniture \\
murky & cloud, mud, pond & tire, wheel, nut \\
old & building, roots, tree & sandwich, bubble, chocolate \\
open & door, window, gate & bus, cliff, granite \\
peeled & orange, potato, fruit & coffee, tea, tower \\
thin & paper, blade, paste & garage, car, garden \\
tiny & penny, toy, town & gear, beef, field \\
torn & fabric, paper, clothes & pond, bronze, clock \\
unripe & fruit, pear, orange & mirror, cat, steel \\
viscous & foam, mud, paste & garden, clock, city \\\hline
    \end{tabular}
    }
    \vspace{1pt}
    \caption{Examples of top-3 and bottom-3 objects associated to 25 randomly selected states, according to ConceptNet feasibility scores.}
    \label{tab:supp-most-feasible-state}
\end{table} 

\section{Additional Quantitative Experiments}
In this section, we report additional experimental results, not included in the main paper due to the lack of space. First, while in the main paper we focused our ablation studies on MIT-States, here we provide additional results also for UT-Zappos and C-GQA. Moreover, we report pCZSL results {when} applying a bias over the seen compositions, for a direct comparison with OW-CZSL results. %

\myparagraph{Ablations on other benchmarks.} Due to space constraints, in the main paper we performed the ablation studies only on MIT-States, following previous works \cite{cge,compcos,symnet,tmn}. However, our results are consistent across settings. As examples, here we also provide ablation studies of the benefit of marginalization on the UT-Zappos dataset in Tab~\ref{tab:supp-marginalization}. Again, we see that marginalization consistently brings improvements to CGE. 
We also evaluate the benefit of hard masking on the C-GQA dataset in Tab.~\ref{tab:supp-hardmask}. We again see hard masking bringing improvements to the HM and AUC on the validation set of C-GQA for both VisProd and KG-SP$_\text{ff}$.

\myparagraph{\settingNickNS: results with bias.} %
In the main paper, we evaluate the models on the \settingNick without applying a bias on the seen compositions. This is because, in the partial setting, we do not have the explicit notion of seen or unseen compositions, thus we cannot follow the same evaluation protocol of standard CZSL and OW-CZSL. Here we show the results on pCZSL when assuming the seen compositions to be known, and applying a bias to them. We underline that this is not fair in the pCZSL setting, but allows us to directly compare results of pCZSL and OW-CZSL. %
 The results are shown in Tab.~\ref{tab:sota-nobias} for all 3 datasets. %
 Consistently with the results without bias, %
 we find that KG-SP always outperforms CompCos and CGE. Specifically, KG-SP achieves significantly better AUC, best harmonic mean, and either comparable or superior best accuracy on seen and unseen compositions. %
 More importantly, the drop from the open-world setting is quite small, \eg 0.78 to 0.61 AUC on C-GQA. This is remarkable, since in pCZSL we have %
 half the labels of OW-CZSL, and indicates the robustness of KG-SP to the available annotations.

\setlength{\tabcolsep}{3pt}
\renewcommand{\arraystretch}{1.2}
\begin{table}[t]
    \centering
    \begin{tabular}{l| c | c c c c }
       & Marginaliz.   &Seen   & Unseen    & HM & AUC    \\\hline
        \multirow{2}{*}{CGE\textsubscript{ff}} && 51.3 & 30.0&25.2 & 11.5 \\
        &\cmark& \textbf{51.4} & \textbf{46.6}& \textbf{30.0} & \textbf{15.4}\\\hline
                  
        \multirow{2}{*}{CGE}  && 51.1  & 47.6 & \textbf{33.3} & 17.5 \\
         &\cmark& \textbf{53.9}& \textbf{48.5} & 32.3 & \textbf{18.1}\\\hline
                  
    \end{tabular}
    \vspace{1pt}
    \caption{OW-CZSL results in the %
    validation set of UT-Zappos when using marginalization. {CGE\textsubscript{ff} is the approach of \cite{cge} with frozen backbone whereas CGE performs end-to-end training.} }%
    \vspace{-10pt}
    \label{tab:supp-marginalization}
\end{table}

\setlength{\tabcolsep}{3pt}
\renewcommand{\arraystretch}{1.2}
\begin{table}[t]
    \centering
    \begin{tabular}{l| c | c c c c }
       & Hard Masking   &Seen   & Unseen    & HM & AUC    \\\hline
        \multirow{2}{*}{VisProd} && 24.8 & 14.8 & 13.2 & 2.8 \\
        &\cmark& \textbf{24.9} & \textbf{14.9} & \textbf{13.3} & \textbf{2.9}\\\hline

        \multirow{2}{*}{KG-SP} && 30.5 & 16.9 & 15.4 & 3.9 \\
        &\cmark& \textbf{30.6} & \textbf{17.0} & \textbf{15.5} & \textbf{4.0}\\\hline

    \end{tabular}
    \vspace{1pt}
    \caption{OW-CZSL results in the %
    validation set of C-GQA when using hard masking. }%
    \vspace{-10pt}
    \label{tab:supp-hardmask}
\end{table}
{
\setlength{\tabcolsep}{6pt}
\renewcommand{\arraystretch}{1.1}
\begin{table*}[t]
    \centering
    \resizebox{0.9\linewidth}{!}{\begin{tabular}{c  l| c c c c | c c c c |  c c c c }
    \multicolumn{2}{c}{\multirow{2}{*}{\textbf{Method}}} & \multicolumn{4}{c}{\textbf{MIT-States}}& \multicolumn{4}{c}{\textbf{UT Zappos}}& \multicolumn{4}{c}{\textbf{C-GQA}}\\
                                                &   
                                                &S   & U    & HM    & AUC   
                                                &S   & U   & HM    & AUC 
                                                &S   & U    & HM    & AUC \\\hline
      &CompCos~\cite{compcos}
      &  18.2        &  \textbf{6.3}     &   5.6 &   0.64  
     &     56.0  & 42.7           & 34.0       &  18.4
     & 22.0  &    1.8   & 2.9   & 0.31\\
    &CGE~\cite{cge}     
     & 19.2   &5.5       & 5.2    & 0.61 
     &  \textbf{60.4} &  \textbf{43.4}     & 34.7   & 19.5
     & 26.7 &  1.2     & 2.1 & 0.25\\\hline
      &KG-SP  
      & \textbf{20.5}   & \textbf{6.3}       & \textbf{5.9}   & \textbf{0.77}     
     & 60.0  &   43.3   & \textbf{40.2}  & \textbf{22.6} 
     & \textbf{29.2} & \textbf{2.4} &\textbf{ 4.1}   &\textbf{0.61}\\\hline
    \end{tabular}}
    \vspace{1pt}
    \caption{\textbf{pCZSL results} on MIT-States, UT Zappos and C-GQA. We measure best seen (S) and unseen accuracy (U), best harmonic mean (HM), and area under the curve (AUC) on the compositions. %
    }
    \vspace{-5pt}
    \label{tab:sota-nobias}
\end{table*}
}

\section{CZSL vs OW-CZSL: an example}
In the main text we described the OW-CZSL setting, proposed in \cite{compcos}. Here, we provide an example to clarify how this setup differs from the more standard CZSL. For more details on this setting, please refer to \cite{compcos}.

Let us consider a toy benchmark where the training set contains images of the following compositions: \textit{wet cat}, \textit{dry apple}, \textit{dry dog}, \textit{ripe apple}. Similarly, let us assume that the same benchmark contains images of the following unseen compositions in the test set: \textit{wet dog}, \textit{dry cat}. Note that other three compositions, \ie \textit{wet apple}, \textit{ripe dog}, \textit{ripe cat}, are not present in any image of the dataset, either because they are unfeasible (\eg \textit{ripe dog}) or because no image has been collected for them (\ie \textit{wet apple}).

The main difference between CZSL and OW-CZSL is that, in the latter, we have no priors on unseen compositions, and we thus consider the full compositional space at test time.  To clarify, in standard CZSL, a model assumes to know which unseen compositions are present in the test set of the dataset and which are not. Thus, a CZSL model would predict 6 compositions: the 4 seen ones during training, and the 2 unseen ones that have at least one image in the test set (\ie \textit{wet dog}, and \textit{dry cat}).  

In OW-CZSL, we do not know which compositions are present in the test set. As a consequence, the output space needs to consider all possible unseen compositions. In our toy benchmark, this means predicting the 4 seen compositions and all the 5 compositions for which we did not have training images (\ie \textit{wet dog}, \textit{dry cat}, \textit{wet apple}, \textit{ripe dog}, \textit{ripe cat}). Note that in OW-CZSL a model needs to cope with the presence of "distractors", \ie compositions close to other existing ones but not present in the dataset (\eg \textit{wet apple} vs \textit{ripe apple}) as well as modeling unfeasible compositions (\eg \textit{ripe dog}) to simplify the task.  While this is a toy example, the difference between the settings is huge in CZSL benchmarks, where these challenges are more pronounced. As an example, CZSL models on MIT-States consider only 1'662 compositions out of the possible 28'175 compositions considered in the output space of OW-CZSL.

As a final note, despite the difference in the output spaces, models built for standard CZSL may perform well in OW-CZSL since they can still exploit what learned from seen compositions to generalize to unseen ones. This can be seen in Table 1 of the main paper, where CGE \cite{cge}, designed for standard CZSL, is competitive in most benchmarks, being either third or second best performing model.

\section{Dataset Licenses}
The datasets used in our work are: UT-Zappos~\cite{utzappos1,utzappos2}, MIT-States~\cite{mitstates}, and C-GQA~\cite{cge} and their licenses are summarized in Table~\ref{tab:licence}. For MIT-States we did not find an accompanying license, but the dataset is publicly available\footnote{\url{http://web.mit.edu/phillipi/Public/states_and_transformations/index.html}}.

\begin{table}[t]
	 \centering
		 \begin{tabular}{ll} 
		 \toprule
		 Dataset & Licence \\
		 \midrule
		 UT-Zappos               & Custom: Non-commercial Usage \\
		 MIT-States               &  Not Available \\
		 C-GQA          & Creative Commons Attribution 4.0 License \\
		 \bottomrule
		 \end{tabular}
	 \caption{The datasets employed in the paper and their licences.}
	 \label{tab:licence}
\end{table}

\end{document}